# Automatic Classification and Segmentation of Tunnel Cracks Based on Deep Learning and Visual Explanations

Yong Feng[1] Xiaolei Zhang[2] Shijin Feng[1,2] Yong Zhao[2] Yihan Chen[1]

(*1. Urban Mobility Institute, Tongji University, 4800 Cao an Rd, Shanghai, 201804, China; 2. Key Laboratory of Geotechnical and Underground Engineering of Ministry of Education, Department of Geotechnical Engineering, Tongji University, 1239 Si ping Rd, Shanghai, 200092, China*)

*E-mail: fengyongtj@tongji.edu.cn*

**Abstract:** Tunnel lining crack is a crucial indicator of tunnels' safety status. Aiming to classify and segment tunnel cracks with enhanced accuracy and efficiency, this study proposes a two-step deep learning-based method. An automatic tunnel image classification model is developed using the DenseNet-169 in the first step. The proposed crack segmentation model in the second step is based on the DeepLabV3+, whose internal logic is evaluated via a score-weighted visual explanation technique. Proposed method combines tunnel image classification and segmentation together, so that the selected images containing cracks from the first step are segmented in the second step to improve the detection accuracy and efficiency. The superior performances of the two-step method are validated by experiments. The results show that the accuracy and frames per second (FPS) of the tunnel crack classification model are 92.23% and 39.80, respectively, which are higher than other convolutional neural networks (CNN) based and Transformer based models. Also, the intersection over union (IoU) and F1 score of the tunnel crack segmentation model are 57.01% and 67.44%, respectively, outperforming other state-of-the-art models. Moreover, the provided visual explanations in this study are conducive to understanding the "black box" of deep learning-based models. The developed two-stage deep learning-based method integrating visual explanations provides a basis for fast and accurate quantitative assessment of tunnel health status.

## 1 Introduction

Due to the issues of traffic congestion and increased population, numerous transportation infrastructures have been constructed and operated during the past decades (Jiang et al., 2023; Feng et al., 2023). With advantages such as high construction efficiency, high flexibility, low resource consumption, and minor environmental disturbance, tunnels have been widely used and become the core infrastructures (Zhang et al., 2023). However, the tunnels in service, as geotechnical infrastructures, are inevitably subjected to the joint action of adjacent excavation disturbance, untimely maintenance, deterioration of materials, temperature variation, improper construction, and groundwater (Feng et al., 2023). Various defects, such as crack, leakage, and spalling, often occur on the tunnel linings, necessitating timely inspection and management.

Numerous statistical results show that lining crack is one of the most severe defects in tunnel engineering (Huang et al., 2022). The occurrence of cracks on the lining surface will affect the reliability and integrity of the tunnel structure to an extent, thereby causing water leakage, concrete corrosion, and reduced lining-bearing capacity of tunnels. Without limiting their expansion, cracks may lead to serious safety accidents (Zhou et al., 2023). Hence, it is necessary to carry out regular crack inspections to ensure the long-term stability and safety of the tunnels. Traditionally, tunnel crack detection relied on manpower. However, the necessity of active use of intelligent recognition technology for tunnel lining cracks has been emphasized due to the low accuracy, low efficiency, high subjectivity, and high risk of manual visual inspections. Deep learning, especially convolutional neural network (CNN), which has made a paradigm shift recently, provides unprecedented opportunities to facilitate fast, accurate, and automatic recognition of tunnel lining cracks.

In the past, many researchers have delved into CNN-based approaches for tunnel lining crack identification. Huang et al. (2018) established a two-stream algorithm based on FCN, where one stream was employed to segment the cracks by sliding-window-assembling operation, and the other was used to segment leakages by resizing-interpolation operation. Yang et al. (2018) also adopted FCN to semantically identify and segment pixel-wise cracks at different scales, and then utilized traditional digital image processing techniques to quantitatively measure the morphological features of cracks. Miao et al. (2019) proposed a novel semantic segmentation by integrating the UNet with squeeze-and-excitation and residual learning blocks to segment crack and spalling defects. Following the same design paradigm, Hou et al. (2021) and Dang et al. (2022) also improved the UNet through adding residual learning units to complete the tunnel lining crack segmentation task. In the work of Gao et al. (2019), an optimal adaptive selection model (RetinaNet-AOS) based on RetinaNet was developed for semantic segmentation on tunnel lining crack images. Xu and Yang (2019) implemented crack identification by means of Mask R-CNN, which has been widely used in the field of tunnel lining defect detection. Similar to the tasks completed by the Mask R-CNN, Zhao et al. (2021) presented an improved PANet model to obtain refined crack segmentation results, achieving complete separation of cracks from the lining backgrounds. After that, Zhou et al. (2022) applied YOLOv4 enhanced by EfficientNet and depthwise separable convolution to detect three types of tunnel lining defects, i.e., crack, water leakage, and rebar-exposed. Considering the excellent detection performance of the YOLO series models, Liu et al. (2022) coupled YOLOv5 with a transfer learning technique to localize the cracks in the road tunnel lining images.

Based on the above, it can be summarized that existing studies have achieved outstanding results in tunnel lining defect identification by investigating more perfect algorithms or improving existing ones. However, in real-world applications, it is common that image acquisition devices acquire a large amount of lining image data, including abundant normal (defect-free) images and anomalous (crack) ones. The existing engineering requirement is to efficiently, accurately, and automatically identify cracks from massive amounts of tunnel image data, which has become a considerable and urgent challenge for previous research. Most of the available studies directly treat the crack recognition problem as a single-attribute classification or separately strive to find the locations or areas of the cracks as a localization or segmentation problem. To address these limitations, this study proposes a tunnel lining crack recognition framework in a two-stage manner, which combines image classification and segmentation. The purpose of tunnel lining image classification is to determine whether the image contains cracks, while the aim of tunnel lining crack segmentation is to extract cracks from backgrounds in crack images, which are different in definition. Image classification is much easier than crack segmentation. Specifically, the image classification task classifies the tunnel lining images as defect-free or crack ones, while the segmentation task getting the crack pixels from images is much more complex and time-consuming. Consequently, a tunnel lining image classification model named DenseNet-169 is adopted in the first stage to classify and save the images containing cracks. Another tunnel lining crack segmentation model called DeepLabV3+ is exploited in the second stage to process the crack images and isolate the cracks from backgrounds.

Nevertheless, deep learning models also have deficiencies and limitations, such as being considered as 'black box' models, which have trouble in explaining physically and rely heavily on hyperparameter settings. To account for mechanisms behind the 'black box' of the crack segmentation model and build trust in deep learning models, this study employs an advanced visual interpretation technology, i.e., score-weighted class activation mapping (Score CAM) (Wang et al., 2020). The motivation behind Score CAM is to generate heatmaps highlighting meaningful regions to intuitively explain the CNN-based models' internal mechanism.

# 2 Methodology

## 2.1 Framework for tunnel lining crack identification

This study designs the tunnel lining crack recognition framework in a two-stage manner, which is shown in Fig. 1. In the beginning, the tunnel lining images collected from practical tunnel engineering are fed into the automatic classification stage, where an image classification model is responsible for selecting and saving crack images from massive tunnel lining image data. Specifically, the DenseNet-169 model takes the lining image as an input and outputs two probabilities to determine whether the image belongs to a crack or defect-free image. Subsequently, the images containing cracks are further input into the second stage to isolate the cracks from the backgrounds. That is, the DeepLabV3+ model is dedicated to identifying the pixels belonging to cracks in the images. Through these two stages, cracks' locations and geometric shapes are extracted with high precision and efficiency. The deep learning models employed in each stage are described in detail in the following subsections. Moreover, a CNN explanation step is also incorporated into the overall framework to understand the operational mechanism of the segmentation model and build trust in CNN.

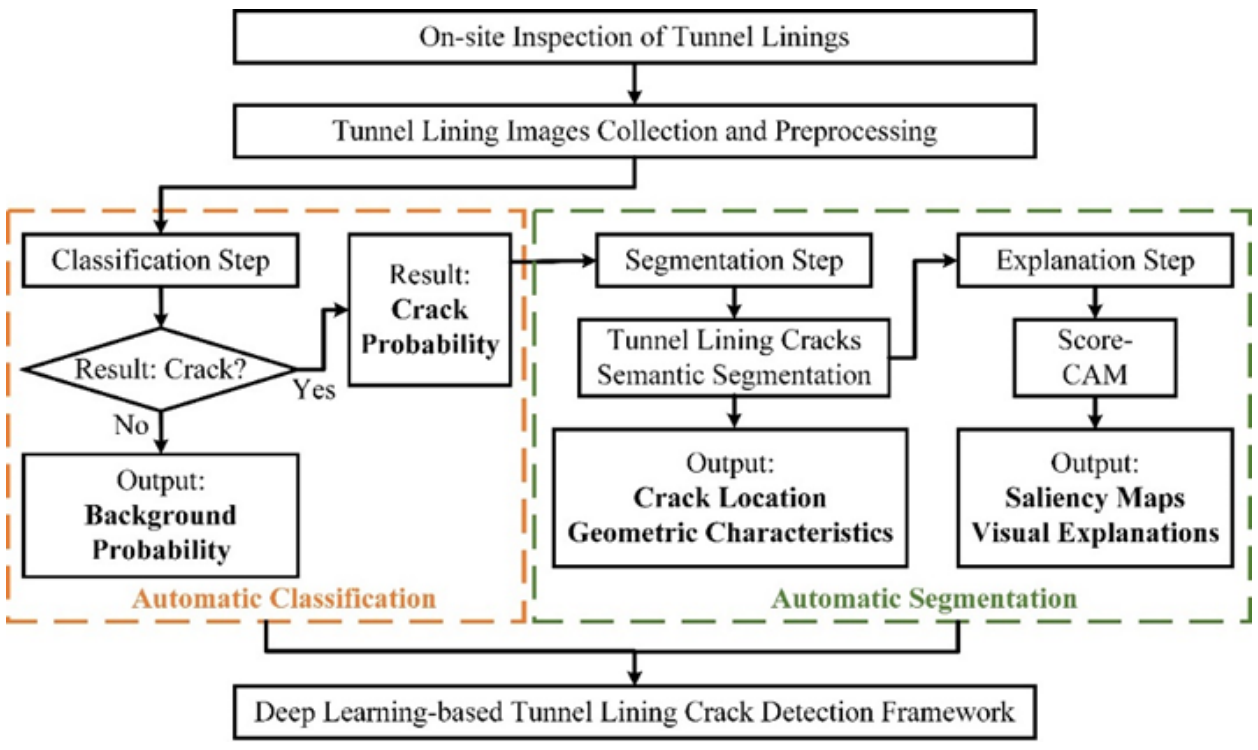

Figure 1 Pipeline of tunnel lining cracks identification

## 2.2 Tunnel lining image classification model

Powered by the successful application of dense convolutional network (DenseNet) (Huang et al., 2018), this study employs DenseNet-169 as the automatic classification model for tunnel lining images in the first step. DenseNet is a type of convolutional neural network that directly connects all layers (with matching feature-map sizes) with each other for maximum information flow between layers in the network. In particular, each layer obtains additional inputs from all preceding layers and passes on its feature maps to all subsequent layers, as observed in Figs. 2 and 3.

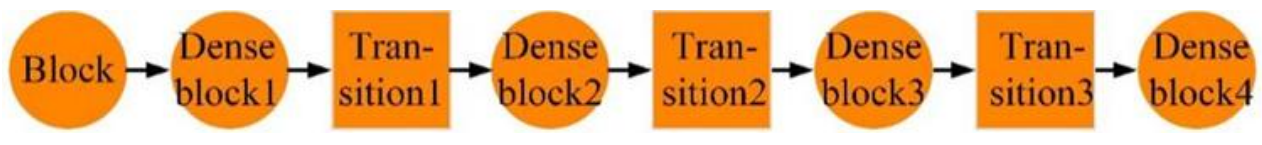

Figure 2 Architecture of DenseNet

DenseNet is a multi-stage architecture where the first stage consists of a standard 7×7 convolution, batch normalization, ReLU activation function, and maximum pooling, followed by four stages consisting of Denseblocks and three stages composed of Transition blocks, and finally, a global average pooling layer and a fully connected layer responsible for outputting the predictions. DenseNet-169 is one of the DenseNets, containing 6, 12, 32, and 32 Denseblocks in four stages, respectively. It is worth pointing out that the cross entropy function is chosen as the model's loss function.

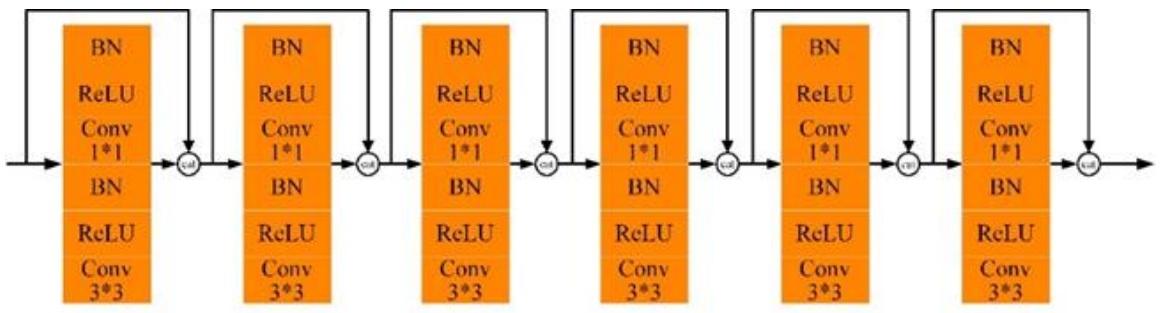

(a) Denseblock

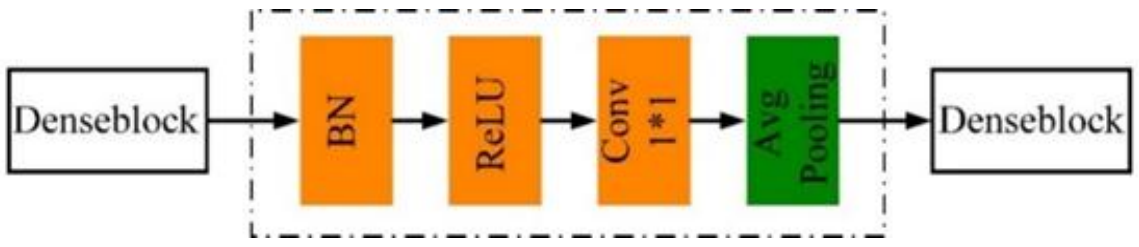

(b) Transition block

Figure 3 Details of modules in DenseNet

## 2.3 Tunnel lining crack segmentation model

After selecting images containing cracks from a large number of tunnel lining images, a crucial task is to segment the cracks from the images in an end-to-end and pixel-to-pixel manner. To this end, DeepLabV3+ (Chen et al., 2018) serves as the crack segmentation model in the second step. Like most semantic segmentation models, the DeepLabV3+ model comprises an encoder and a decoder, with the overall architecture shown in Fig. 4.

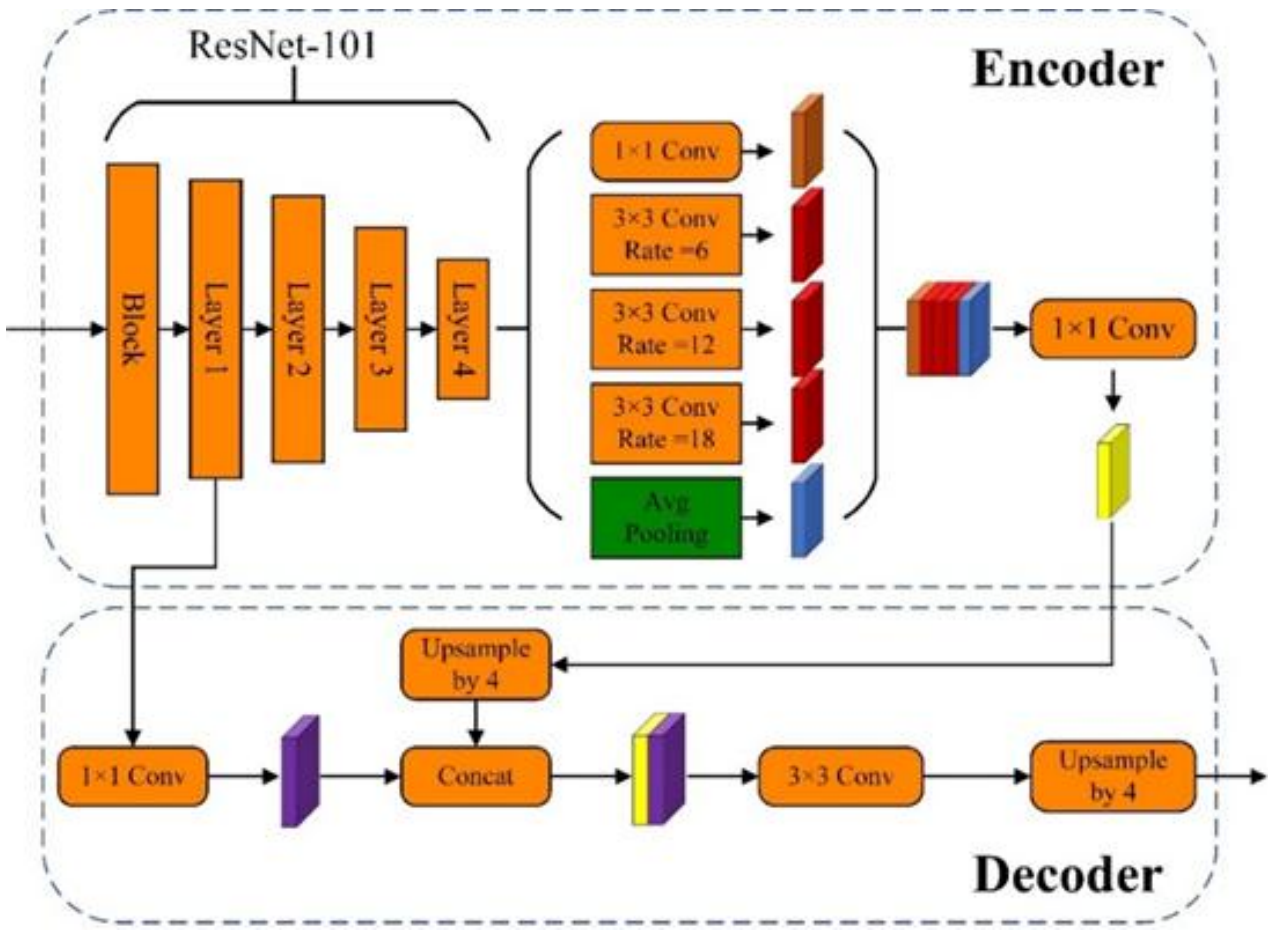

Figure 4 Overall architecture of the DeepLabV3+ model

ResNet-101 (He et al., 2016) is used as the backbone structure to extract shallower features containing spatial information and deeper features containing semantic information for tunnel lining cracks. Concretely, the ResNet-101 includes a 'Block' for downsampling operation with a factor of four, followed by 'Layer 1', 'Layer 2', 'Layer 3', and 'Layer 4' for feature extraction. The innovation of the DeepLabV3+ model is the atrous spatial pyramid pooling (ASPP) in the encoder. Discontinuity problem is prevalent in crack segmentation tasks. The reason for this phenomenon is the loss of multi-scale information about the cracks. To compensate for this deficiency, the ASPP structure is created with multiple dilated convolutions. For a dilated convolution with a dilation rate of d and kernel size of N, the receptive field reaches the same as that of a standard convolution with a kernel size of $(N - 1) \times d + 1$. As a result, dilated convolution is capable of increasing the receptive field while maintaining the size of the feature maps, thereby reducing the loss of crack feature information caused by consecutive convolution and pooling downsampling operations. In ASPP, three dilated convolution operations with dilation rates of 6, 12, and 18 conduct multi-scale sampling on the feature maps output by ResNet-101. In addition, the ASPP integrates a 1×1 standard convolution operation and an average pooling operation, which together with the dilated convolutions, form feature maps containing multi-scale information about the cracks. A 1×1 convolution then processes the features to realize further feature extraction and dimensionality reduction. In the decoder, the shallow features generated by 'Layer 1' of ResNet-101 are collected and then concatenated with upsampled deep features obtained from ASPP. Skip connection enables the deep learning model to combine local spatial features of cracks with global context information, thus further relieving the discontinuity problem in crack segmentation. Fused features are then subjected to successive 3×3 convolution and 4× bilinearly upsampling operations to restore the size of the original input image and produce the final semantic segmentation result.

Last but not least, the loss function is also crucial in segmenting tunnel lining cracks. Tunnel lining cracks are slender and small targets in the image compared to the background. To put it differently, the crack pixels are much less than the background pixels in tunnel lining images. Therefore, the imbalanced data will make the segmentation model biased towards the background class rather than cracks. To address this issue, dice loss is selected as the loss function of the DeepLabV3+ model.

## 2.4 Visual explanation methods

As described before, this study aims to understand and explore the mechanisms behind the 'black box' of the crack segmentation model. An advanced visual interpretation technology, i.e., Score CAM, is leveraged, and its detailed processing steps are as follows.

Given the CNN model (i.e., DeepLabV3+ in this study) $Y = f(X)$ that takes $X$ as an input and outputs a scalar result $Y$. We select a targeted convolutional layer $l$ in $f$ and the corresponding activations as $A$. Define the *k-th* channel of $A_l$ as $A_l^k$ . Given a known baseline input $X_b$ , the contribution of $A_l^k$ to $Y$ is expressed as follows.

$$C(A_l^k) = f(X \circ H_l^k) - f(X_b) \quad (1)$$

$$H_l^k = s(Up(A_l^k)) \quad (2)$$

where $Up(\cdot)$ represents the up-sampling operation of $A_l^k$ into the input size. $s(\cdot)$ denotes the normalization operation converting each element in the input matrix to the interval [0, 1].

For a class of interest $c$ (i.e., crack in this study), Score-CAM $L^{c}_{\text{Score}-CAM}$ is defined as follows.

$$L^{c}_{Score-CAM} = ReLU(\sum C(A_l^k) A_l^k) \quad (3)$$

Define a weight coefficient $\alpha_k^c = C(A_l^k)$, then the equation (3) can be abbreviated to equation (4).

$$L^{c}_{(Score-CAM)} = ReLU(\sum \alpha_k^c A_l^k) \quad (4)$$

Based on the above steps, visual heatmaps can be acquired to understand the crack segmentation model's internal structure and decision-making mechanism.

## 3 Data

All experiments in this study were carried out based on an open-source dataset named NUAACrack-2000 (Qiu et al., 2022; Zhang et al., 2021), where the tunnel lining images were captured in China. On the basis of the NUAACrack-2000, two datasets were established: one was exploited for training and testing of the DenseNet-169 model, and the other for training and testing of the DeepLabV3+ model.

Fig. 5 illustrates the typical crack and defect-free images in the image classification dataset, and Table 1 summarizes the number of different types of images. A total of 1942 images with a pixel size of 512×375 were divided into training set, validation set, and testing set according to a ratio of 7:2:1.

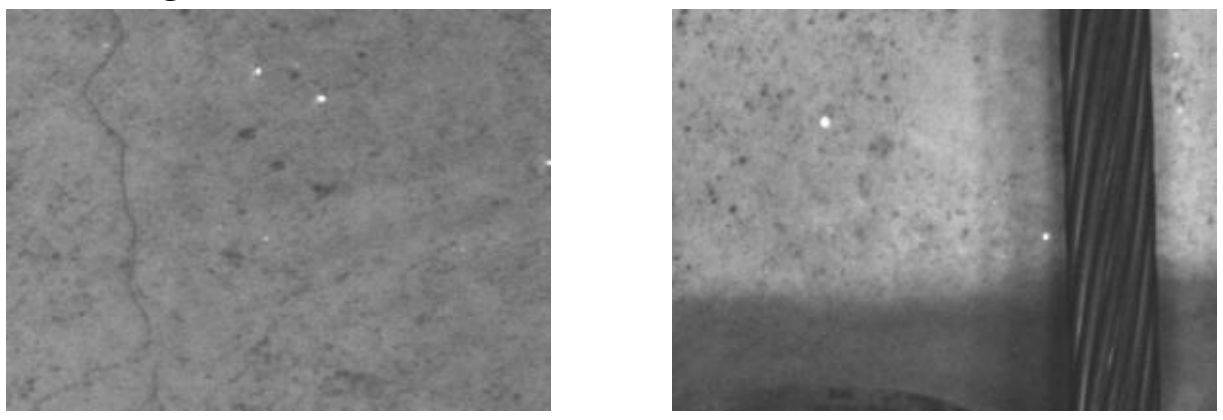

Figure 5 Examples of tunnel lining image classification dataset

**Table 1 Dataset for image classification experiment**

| | **Training** | **Validation** | **Testing** | Total |
|---|---|---|---|---|
| **Background** | 333 | 94 | 47 | 474 |
| **Crack** | 1029 | 293 | 146 | 1468 |
| Total | 1362 | 387 | 193 | 1942 |

After that, 1468 crack images and their ground truths made up the semantic segmentation dataset, as shown in Fig. 6. Each crack image corresponds to a label, which was used to supervise the training of the segmentation model. These crack images were also partitioned based on the holdout approach in a ratio of 7:2:1. That is, 1028 images were used for training, 294 images for verification, and the remaining 146 images for testing.

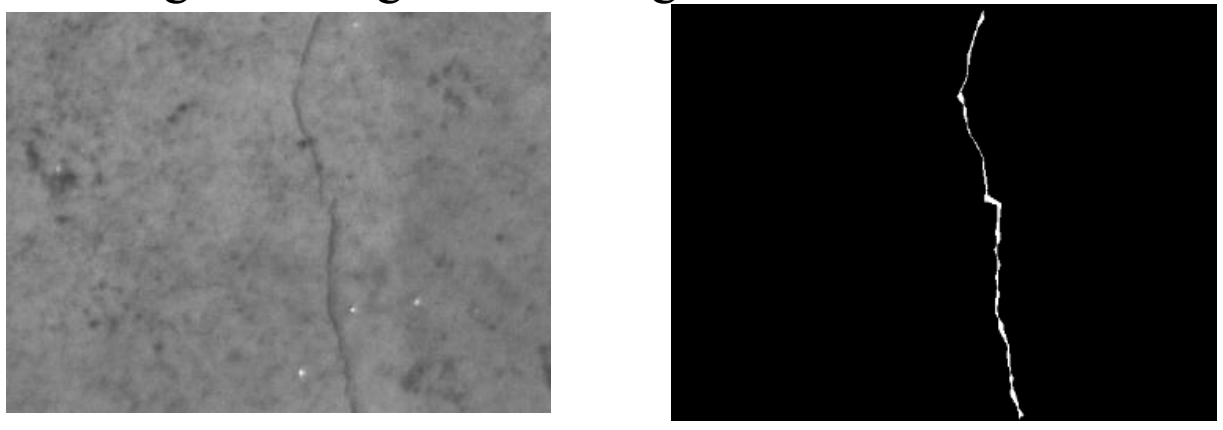

Figure 6 Examples of crack segmentation dataset

## 4 Experiment

The experiments in this study mainly involve two parts: image classification and crack segmentation. In the first part, the image classification dataset was utilized to train and test the DenseNet-169 model. In the second part, the crack segmentation dataset was used for the training and testing of the DeepLabV3+ model. Comparative experiments with other dominant models and visual explanations were also included in this section.

### 4.1 Experimental environment

Experiments were conducted under the computer configurations of the Windows 10 operating system, one NVIDIA GeForce RTX 3090 graphics processing unit (GPU), and one Intel Core i9-12900KF central processing unit (CPU). Python 3.8.12, Pytorch 1.12.0, CUDA 11.6, and CUDNN 8.3 constitute the computation software environment.

### 4.2 Tunnel lining image classification experiment

#### 4.2.1 Evaluation criteria

This study adopted two commonly used measures, accuracy and frames per second (FPS), to evaluate the performance of the classification model.

#### 4.2.2 Model training and testing

To accelerate training speed and save computational resources, all images in the classification dataset were resized to 224×224 before being input into the model. The batch size was set to four; that is, the model processed four images simultaneously in each iteration. To ensure the convergence of the model, the training epoch is determined to be 100. The learning rate is another important hyperparameter. For the purpose of finding the global minimum of the deep learning model, a dynamic learning rate schedule was adopted instead of a constant learning rate. Specifically, the initial learning rate of 0.005 was decreased every ten epochs with an attenuation factor of 0.1. After completing training, the performance evaluation of the model was performed on 193 testing images. Fig. 7 intuitively displays the predicted results of the DenseNet-169 model on several images in the testing set. As seen in Fig. 7, the deep learning model automatically outputs two probability values for each image and then determines the image's category. The first row in Fig. 7 lists four images classified as background, and the second row shows the examples identified as crack images. Through calculation, the accuracy of the DenseNet-169 model reaches 92.23%, and the FPS reaches 39.80.

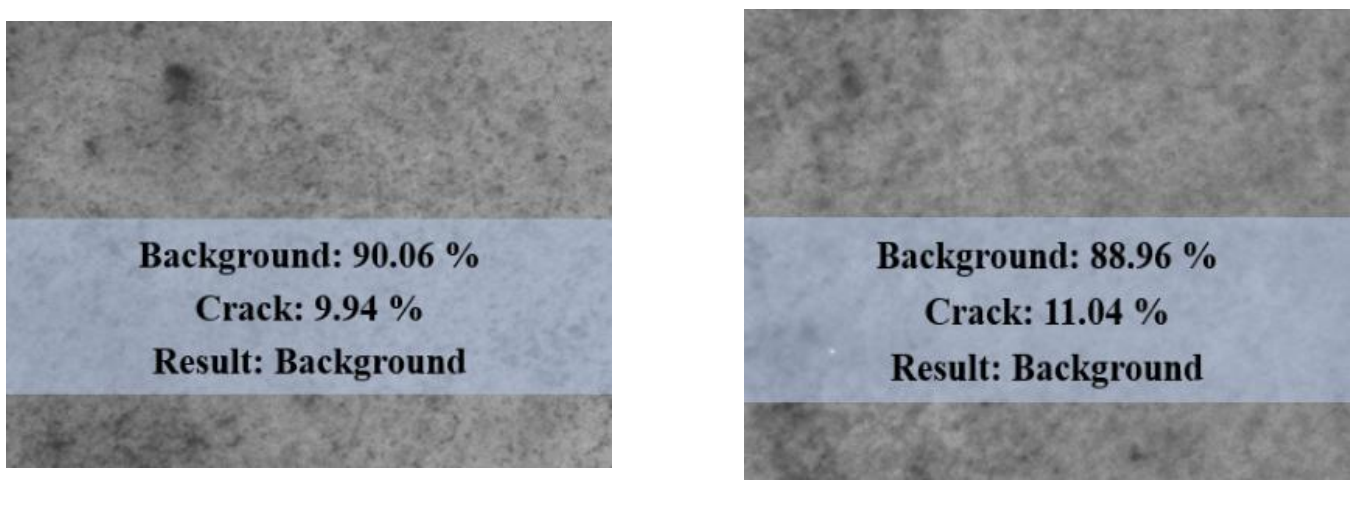

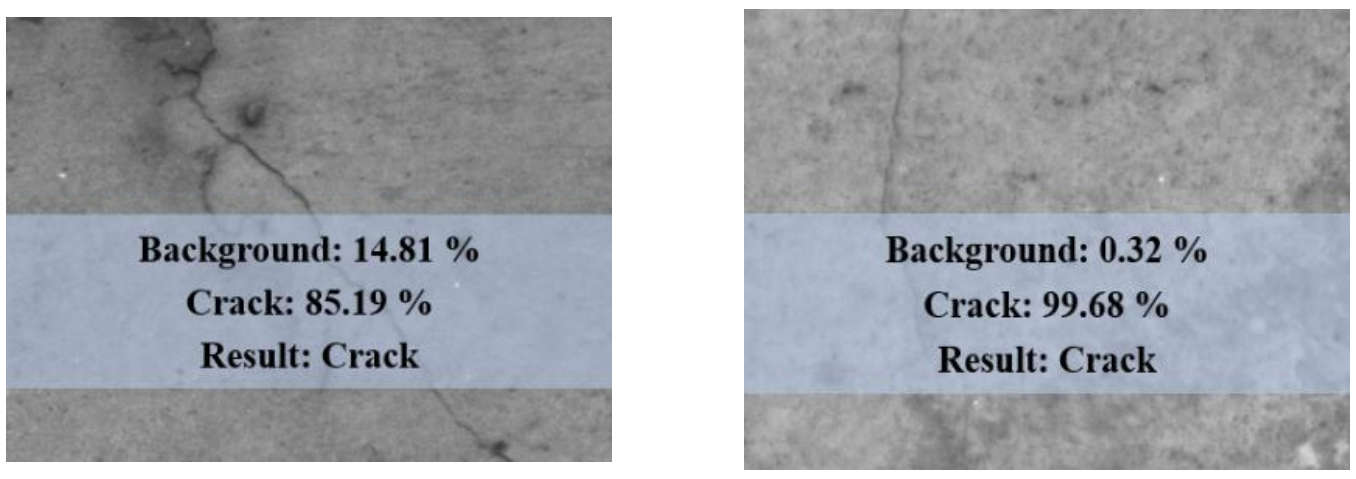

Figure 7 Predictions of DenseNet-169 on several testing images

4.2.3 Comparison with other models

For the sake of further evaluating the classification performance of the DenseNet-169 model for tunnel lining images, six state-of-the-art deep learning models were implemented for comparison. The models used for comparison include CNN-based models and visual Transformer-based models, namely DenseNet-201, EfficientNet-B0, ResNet-50, ResNet-101, Swin Transformer, and Vision Transformer. As reported in Table 2, the DenseNet-169 outperforms the other six deep learning models in terms of accuracy, as reflected in the 2.07%, 16.58%, 17.10%, 18.65%, 16.58%, and 16.58% improvement over DenseNet-201, EfficientNet-B0, ResNet-50, ResNet-101, Swin Transformer, and Vision Transformer.

**Table 2 Results of image classification experiments**

| Models | Accuracy (%) | FPS (f/s) |
|---|---|---|
| **DenseNet-169** | **92.23** | 39.80 |
| **DenseNet-201** | 90.16 | 35.60 |
| **EfficientNet-B0** | 75.65 | 40.77 |
| **ResNet-50** | 75.13 | 60.22 |
| **ResNet-101** | 73.58 | 51.68 |
| **Swin Transformer** | 75.65 | 42.48 |
| **Vision Transformer** | 75.65 | **69.64** |

DenseNet-169 incorrectly classified 12 crack images as defect-free images. As for Swin Transformer and Vision Transformer, these two models wrongly classified all defect-free images as crack images, resulting in the lowest accuracy. The reason for this is inferred to be twofold: on the one hand, the small size of the classification dataset used in this paper prevents the self-attention mechanism in the visual Transformer from realizing its potential, and on the other hand, the imbalance in the number of positive and negative samples leads to a bias in the recognition ability of the models towards the crack images.

From the perspective of the running speed of the models, the FPS of DenseNet-169 is higher than that of DenseNet-201 but lower than that of other models. Improving the running speed of the classification model deserves further research in the future.

### 4.3 Tunnel lining crack segmentation experiment

4.3.1 Evaluation criteria

This study exploited four metrics widely used in semantic segmentation tasks, i.e., precision, recall, F1 score, and Intersection over Union (IoU), to comprehensively assess the crack segmentation models.

IoU and threshold d were introduced to compare the similarity between two arbitrary cracks. For crack images, the IoU between the segmented crack and ground truth was calculated to judge whether the crack has been detected. Generally, the threshold d was set to 0.5 (Liu and Wang, 2022). In this study, when the IoU is greater than 0.5, it is considered that the model has detected out the crack; otherwise, it is not. IoU also served as the primary evaluation indicator for segmentation models in this paper.

4.3.2 Model training and testing

During the training process, the original images were first resized to 512×384 pixels, which would not cause a large amount of calculation and is easy to work with. Accordingly, the batch size was set as 8. The learning rate was also dynamically adjusted, with a value of 0.001 for the first 50 epochs and 0.0001 for the last 50 epochs. A total of 100 rounds of training ensured the convergence of the crack segmentation models. After each round of training, the model was evaluated on the validation set, and the training and validation loss curves were recorded in Fig. 8. It can be seen from Fig. 8 that the training loss declines rapidly over the first ten epochs, then slowly decreases in the next 40 epochs, and stabilizes around 0.2 over the last 50 epochs. Validation loss shows fluctuations in the first 50 epochs and stabilizes in the next 50. Based on the values of the loss function on the validation set, the optimal trained model was determined and saved accordingly.

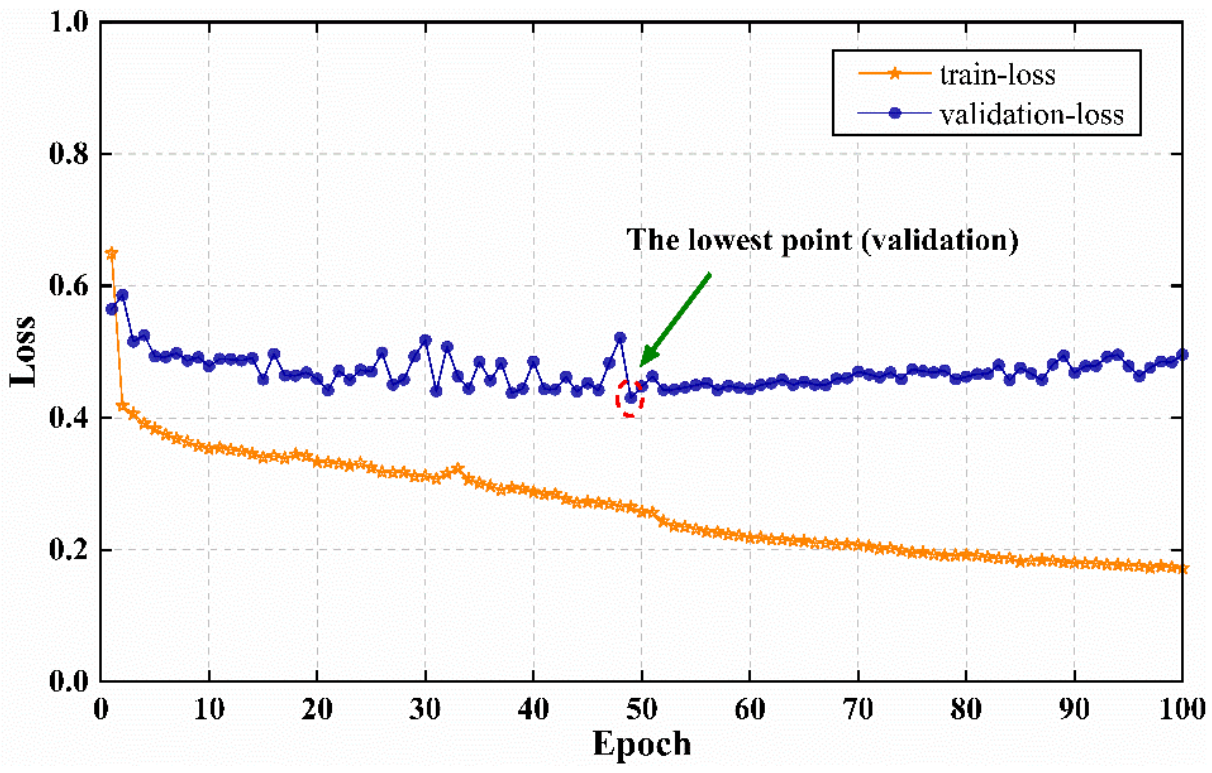

Figure 8 Visualization diagram of loss curves of DeepLabV3+ model during training and validation process

The best-trained model was tested using crack images in the semantic segmentation dataset, and the results showed that the DeepLabV3+ model achieved 57.01% IoU, 67.44% $F_1$ score, 62.95% precision, and 82.12% recall. There is a significant gap between the precision and recall of the DeepLabV3+ model. It should be emphasized that deep learning models must detect all cracks as much as possible from the perspective of engineering applications. Hence, it is reasonable to improve the recall of tunnel lining crack segmentation during practical inspection tasks, even at the expense of precision (Ren et al., 2020).

4.3.3 Comparison with other models

Extensive comparative tests were completed to further verify the applicability and superiority of the DeepLabV3+ model in tunnel lining crack segmentation. Four dominant semantic segmentation models, namely DeepLabV3 with backbone ResNet-101, PSPNet with backbone ResNet-50, UNet backboned by VGG-13, and UNet++ backboned by VGG-13 were trained with the crack segmentation dataset for more comprehensive comparisons.

As shown in Table 3, the DeepLabV3+ model obtains the best IoU of 57.01%, which is 1.77%, 5.77%, 0.17%, and 0.59% higher than DeepLabV3, PSPNet, UNet, and UNet++, respectively. Similarly, the precision of DeepLabV3+ is the highest, improved by 2.02%, 3.72%, 0.60%, and 2.64% compared to DeepLabV3, PSPNet, UNet, and UNet++, respectively. The F1 score for DeepLabV3+ is only slightly lower than UNet but higher than the other three models. The recall of the DeepLabV3+ model is lower than that of UNet and UNet++ but 0.83% and 8.52% higher than that of DeepLabV3 and PSPNet, respectively.

**Table 3 Results of semantic segmentation experiments**

| Models | IoU (%) | $F_1$ score (%) | Precision (%) | Recall (%) |
|---|---|---|---|---|
| **DeepLabV3+** | **57.01** | 67.44 | **62.95** | 82.12 |
| **DeepLabV3** | 55.24 | 66.33 | 60.93 | 81.29 |
| **PSPNet** | 51.24 | 62.18 | 59.23 | 73.60 |
| **UNet** | 56.84 | **67.68** | 62.35 | 83.91 |
| **UNet++** | 56.42 | 67.31 | 60.31 | **84.91** |

Representative testing results predicted by different deep learning models are depicted in Fig. 9 to complement the quantitative comparison results above, with the orange dashed box and red dashed box denoting the incorrect and missed detections. From top to bottom of Fig. 9, they are the raw images in the crack segmentation dataset, labels, and predicted results made by DeepLabV3+, DeepLabV3, PSPNet, UNet++, and UNet. The DeepLabV3+ model performs best on the four testing samples, as seen by the comparison of segmentation results of different models. For case 1, the cracks can be accurately isolated by DeepLabV3+ and PSPNet. However, a small section of crack is omitted by DeepLabV3. There are some false detections in the identification results of UNet. UNet++ also has the situation of false detections. For case 2, all models successfully segment the cracks, except for PSPNet, ignoring large sections of the crack. As seen in the third and fourth columns of Fig. 9, the features of the cracks do not differ significantly from the background features, posing a challenge for crack identification. In case 3, a small portion of cracks are not segmented by PSPNet and UNet, and a small background area is falsely identified as cracks by UNet++. In case 4, DeepLabV3, PSPNet, UNet++, and UNet have different degrees of missed detections. Additionally, a small portion of background pixels are mistakenly detected as crack pixels. The width of the crack segmented by the DeepLabV3+ model is larger than the actual crack width. Nevertheless, the predicted result retains more edge information relative to the other four models.

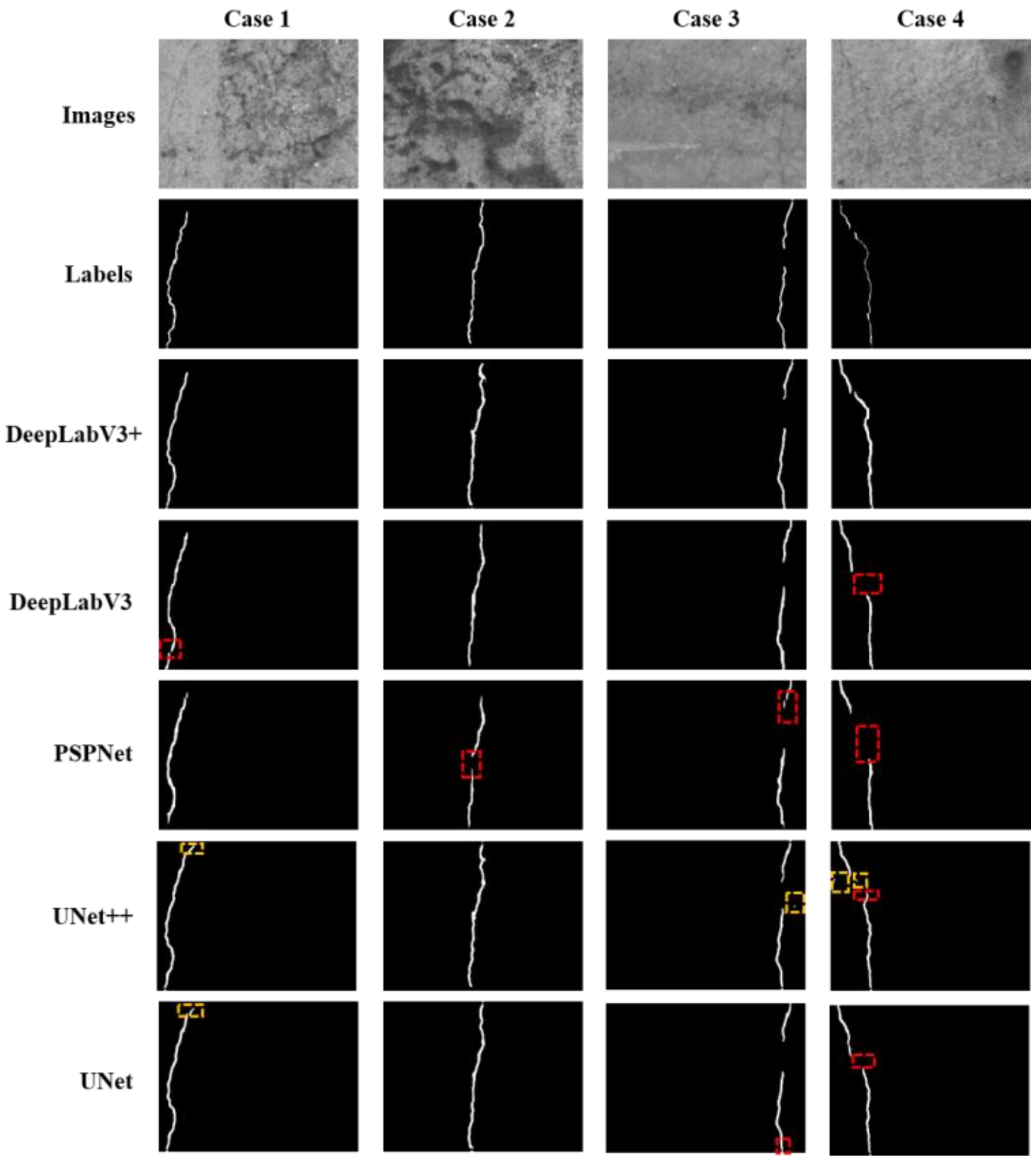

Figure 9 Comparison of prediction performance of different models on the test set

These quantitative and qualitative comparative results above indicate that adopting the DeepLabV3+ model in the second step of our proposed framework for crack segmentation is effective and superior.

### 4.4 Visual explanations

The heatmaps of some key modules of the DeepLabv3+ model are generated based on the Score CAM technique to focus on the encoding and decoding process of the model and understand the decision-making process. For a brief discussion, Fig. 10 presents the heatmaps of the key modules of the DeepLabV3+ model when a single tunnel crack image is used as input. As depicted in Fig. 10, the model focuses on the image globally during the encoding process. At the stage of Layer 2, the model pays attention to crack, lining surface, and lighting. Until the stage of Layer 4, the model emphasizes the crack, but the highlighted areas (dark-colored) do not precisely fit the crack. During decoding, the DeepLabV3+ model gradually focuses on the crack itself. Another interesting finding is that the low-level and high-level feature maps used for the concatenation operation shown in Fig. 4 present completely different heatmaps. The heatmaps demonstrate that high-level feature maps focus on the global features of the tunnel lining image due to the inclusion of strong semantic information, while low-level feature maps focus on local regions of the image, confirming that shallower feature maps contain rich spatial information.

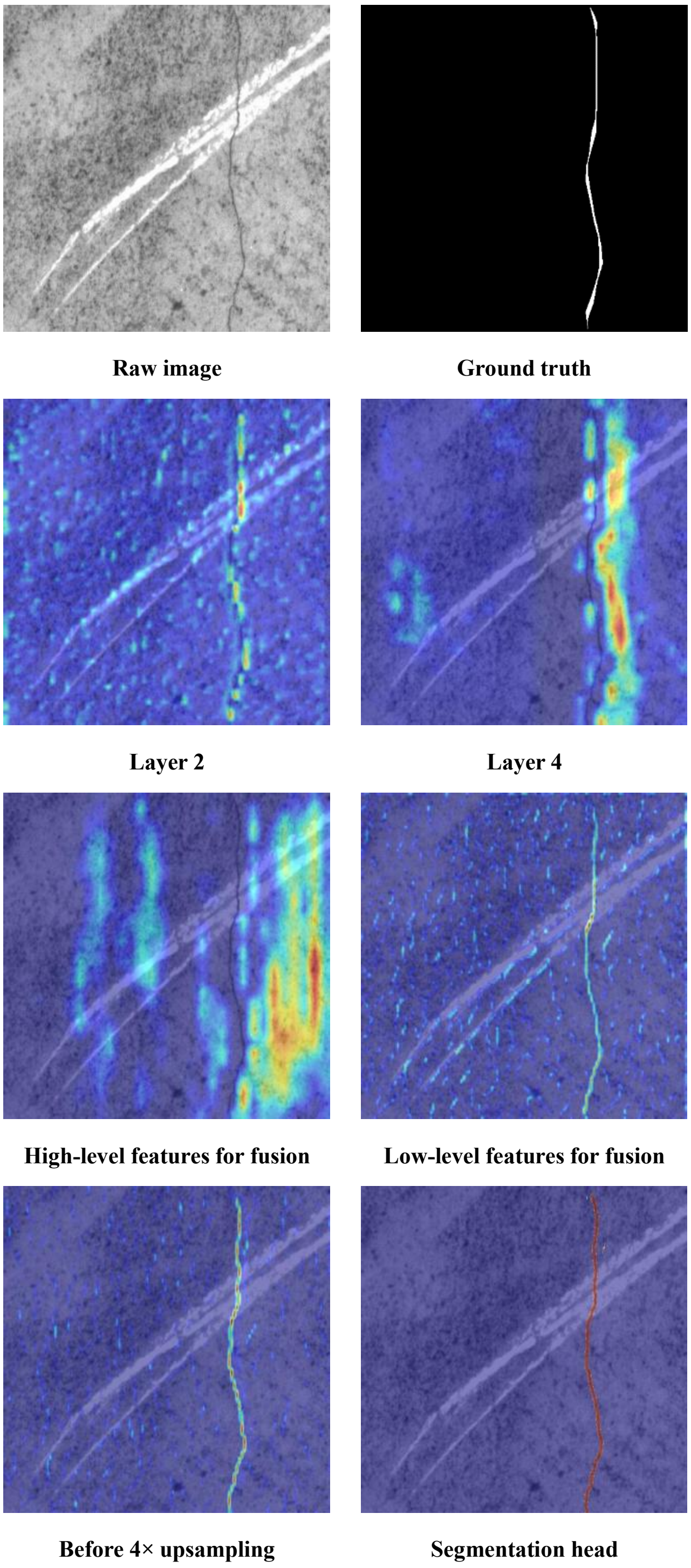

Figure 10 Visual explanations for several key modules of the DeepLabv3+ model based on Score CAM

## 5 Conclusion

This study proposes a two-step deep learning-based method for the automatic classification and segmentation of tunnel cracks. DenseNet-169 serves as the tunnel lining image classification model in the first step, through which the crack images can be separated and saved from massive image data. In the second step, the DeepLabV3+ model is employed to separate cracks from backgrounds in the crack images. An advanced visual explanation technique is integrated into the two-step method to understand the 'black box' of the crack segmentation model. The superiority and rationality of the two-step method are demonstrated through extensive comparative experiments. DenseNet-169 achieves 92.23% accuracy, which is improved by 2.07%, 16.58%, 17.10%, 18.65%, 16.58%, and 16.58% over DenseNet-201, EfficientNet-B0, ResNet-50, ResNet-101, Swin Transformer, and Vision Transformer. The FPS of DenseNet-169 reaches 39.80, exceeding that of DenseNet-201. For DeepLabV3+, an IoU of 57.01% is obtained, which is 1.77%, 5.77%, 0.17%, and 0.59% higher than DeepLabV3, PSPNet, UNet, and UNet++, respectively. Furthermore, the provided visual explanations show that the segmentation model focuses on the image globally during the encoding process and gradually focuses on the crack itself during the decoding process. Another interesting finding is high-level feature maps focus on the global features of the tunnel lining image due to the inclusion of strong semantic information. In contrast, low-level feature maps focus on local regions of the image, confirming that shallower feature maps contain rich spatial information.

There are still some flaws that need to be addressed in the future. Firstly, it is urgently required to introduce model light-weighting techniques to accelerate the image classification models. Secondly, essential image pre-processing or post-processing approaches need to be incorporated into the crack segmentation process to eliminate the complex environmental interference, thus obtaining refined segmentation results. Thirdly, expanding the tunnel lining image classification and crack segmentation datasets to guarantee higher generalization and robustness of the models. Besides, deep learning-based models should be integrated into intelligent detection devices, such as unmanned aerial vehicles (UAVs), to realize automatic and real-time detection of tunnel lining cracks.

## Acknowledgments

Much of the work described in this paper was supported by the National Key Research and Development Program of China under Grant No. 2020YFC1808105, Science and Technology Commission of Shanghai Municipality under Grant No. 21DZ1204400, and the National Natural Science Foundation of China under Grant No. 42372335. The writers would like to greatly acknowledge all these financial supports and express their most sincere gratitude.

## REFERENCES

Chen, L.-C., Zhu, Y., Papandreou, G., Schroff, F., Adam, H., 2018. Encoder-Decoder with Atrous Separable Convolution for Semantic Image Segmentation. http://arxiv.org/abs/1802.02611

Dang, L.M., Wang, H., Li, Y., Park, Y., Oh, C., Nguyen, T.N., Moon, H., 2022. Automatic tunnel lining crack evaluation and measurement using deep learning. Tunnelling and Underground Space Technology 124, 104472. https://doi.org/10.1016/j.tust.2022.104472

Feng, S., Feng, Y., Zhang, X., Chen, Y., 2023. Deep learning with visual explanations for leakage defect segmentation of metro shield tunnel. Tunnelling and Underground Space Technology 136, 105107. https://doi.org/10.1016/j.tust.2023.105107

Feng, Y., Zhang, X., Feng, S., Chen, H., Zhao, Y., Chen, Y., 2023. Improved SOLOv2 detection method for shield tunnel lining water leakages. Journal of Intelligent Construction 1, 9180004. https://doi.org/10.26599/JIC.2023.9180004

Gao, X., Jian, M., Hu, M., Tanniru, M., Li, S., 2019. Faster multi-defect detection system in shield tunnel using combination of FCN and faster RCNN. Advances in Structural Engineering 22, 2907–2921. https://doi.org/10.1177/1369433219849829

He, K., Zhang, X., Ren, S., Sun, J., 2016. Deep Residual Learning for Image Recognition, in: 2016 IEEE Conference on Computer Vision and Pattern Recognition (CVPR). Presented at the 2016 IEEE Conference on Computer Vision and Pattern Recognition (CVPR), IEEE, Las Vegas, NV, USA, pp. 770–778. https://doi.org/10.1109/CVPR.2016.90

Hou, S.K., Ou, Z.G., Qin, P.X., Wang, Y.L., Liu, Y.R., 2021. Image-based crack recognition of tunnel lining using residual U-Net convolutional neural network. IOP Conf. Ser.: Earth Environ. Sci. 861, 072001. https://doi.org/10.1088/1755-1315/861/7/072001

Huang, G., Liu, Z., van der Maaten, L., Weinberger, K.Q., 2018. Densely Connected Convolutional Networks. http://arxiv.org/abs/1608.06993

Huang, H., Li, Q., Zhang, D., 2018. Deep learning based image recognition for crack and leakage defects of metro shield tunnel. Tunnelling and Underground Space Technology 77, 166–176. https://doi.org/10.1016/j.tust.2018.04.002

Huang, H., Zhao, S., Zhang, D., Chen, J., 2022. Deep learning-based instance segmentation of cracks from shield tunnel lining images. Structure and Infrastructure Engineering 18, 183–196. https://doi.org/10.1080/15732479.2020.1838559

Jiang, Y., Wang, L., Zhang, B., Dai, X., Ye, J., Sun, B., Liu, N., Wang, Z., Zhao, Y., 2023. Tunnel lining detection and retrofitting. Automation in Construction 152, 104881. https://doi.org/10.1016/j.autcon.2023.104881

Liu, F., Wang, L., 2022. UNet-based model for crack detection integrating visual explanations. Construction and Building Materials 322, 126265. https://doi.org/10.1016/j.conbuildmat.2021.126265

Liu, J., Zhao, Z., Lv, C., Ding, Y., Chang, H., Xie, Q., 2022. An image enhancement algorithm to improve road tunnel crack transfer detection. Construction and Building Materials 348, 128583. https://doi.org/10.1016/j.conbuildmat.2022.128583

Miao, X., Wang, J., Wang, Z., Sui, Q., Gao, Y., Jiang, P., 2019. Automatic Recognition of Highway Tunnel Defects Based on an Improved U-Net Model. IEEE Sensors J. 19, 11413–11423. https://doi.org/10.1109/JSEN.2019.2934897

Qiu, J., Yan, X., Wang, J., Guo, Y., Wei, M., 2022. Crack extraction from single tunnel image based on fully convolutional neural network. Computer Engineering and Science 44, 845–854. https://doi.org/10.3969/j.issn.1007-130X.2022.05.010

Ren, Y., Huang, J., Hong, Z., Lu, W., Yin, J., Zou, L., Shen, X., 2020. Image-based concrete crack detection in tunnels using deep fully convolutional networks. Construction and Building Materials 234, 117367. https://doi.org/10.1016/j.conbuildmat.2019.117367

Wang, H., Wang, Z., Du, M., Yang, F., Zhang, Z., Ding, S., Mardziel, P., Hu, X., 2020. Score-CAM: Score-Weighted Visual Explanations for Convolutional Neural Networks. http://arxiv.org/abs/1910.01279

Xu, X., Yang, H., 2019. Intelligent crack extraction and analysis for tunnel structures with terrestrial laser scanning measurement. Advances in Mechanical Engineering 11, 168781401987265. https://doi.org/10.1177/1687814019872650

Yang, X., Li, H., Yu, Y., Luo, X., Huang, T., Yang, Xu, 2018. Automatic Pixel-Level Crack Detection and Measurement Using Fully Convolutional Network: Pixel-level crack detection and measurement using FCN. Computer-Aided Civil and Infrastructure Engineering 33, 1090–1109. https://doi.org/10.1111/mice.12412

Zhang, J., Chu, W., Tu, W., Su, H., Lu, S., Xu, Y., 2023. Computer Vision-based Monitoring Method for Differential Settlement of Shield Tunnels. J. Phys.: Conf. Ser. 2519, 012057. https://doi.org/10.1088/1742-6596/2519/1/012057

Zhang, Y., Li, X., Qiu, J., Zhai, X., Wei, M., 2021. GFU-Net: A Deep Learning Approach for Automatic Metal Crack Detection, in: Zhang, H., Yang, Z., Zhang, Z., Wu, Z., Hao, T. (Eds.), Neural Computing for Advanced Applications, Communications in Computer and Information Science. Springer Singapore, Singapore, pp. 375–388. https://doi.org/10.1007/978-981-16-5188-5_27

Zhao, S., Zhang, D., Xue, Y., Zhou, M., Huang, H., 2021. A deep learning-based approach for refined crack evaluation from shield tunnel lining images. Automation in Construction 132, 103934. https://doi.org/10.1016/j.autcon.2021.103934

Zhou, Z., Zhang, J., Gong, C., 2023. Hybrid semantic segmentation for tunnel lining cracks based on Swin Transformer and convolutional neural network. Computer aided Civil Eng mice.13003. https://doi.org/10.1111/mice.13003

Zhou, Z., Zhang, J., Gong, C., 2022. Automatic detection method of tunnel lining multi-defects via an enhanced You Only Look Once network. Computer aided Civil Eng 37, 762–780. https://doi.org/10.1111/mice.12836